\documentclass[a4paper,fleqn]{cas-dc}

\usepackage[authoryear,longnamesfirst]{natbib}

\def\tsc#1{\csdef{#1}{\textsc{\lowercase{#1}}\xspace}}
\tsc{WGM}
\tsc{QE}
\begin{document}
\let\WriteBookmarks\relax
\def\floatpagepagefraction{1}
\def\textpagefraction{.001}

% Short title
\shorttitle{}    

% Short author
\shortauthors{}  

% Main title of the paper
\title [mode = title]{Intent-Context Synergy Reinforcement Learning for Autonomous UAV Decision-Making in Air Combat}  

% Title footnote mark
% eg: \tnotemark[1]
% \tnotemark[1] 

% Title footnote 1.
% eg: \tnotetext[1]{Title footnote text}
\tnotetext[1]{} 

% First author
%
% Options: Use if required
% eg: \author[1,3]{Author Name}[type=editor,
%       style=chinese,
%       auid=000,
%       bioid=1,
%       prefix=Sir,
%       orcid=0000-0000-0000-0000,
%       facebook=<facebook id>,
%       twitter=<twitter id>,
%       linkedin=<linkedin id>,
%       gplus=<gplus id>]

\author[1,2]{Jiahao Fu}%[<options>]

% Footnote of the first author
% \fnmark[1]

% Email id of the first author
\ead{fjh@mail.nwpu.edu.cn}

% URL of the first author
\ead[url]{}

% Credit authorship
% eg: \credit{Conceptualization of this study, Methodology, Software}
\credit{Conceptualization of this study, Methodology, Writing - Original draft preparation}

% Address/affiliation
\affiliation[1]{organization={National Elite Institute of Engineering,Northwestern Polytechnical University},
            addressline={}, 
            city={Xi'an},
%          citysep={}, % Uncomment if no comma needed between city and postcode
            postcode={}, 
            state={Shanxi},
            country={China}}

\author[2]{Feng Yang}%[]
% Corresponding author indication
\cormark[1]
% Footnote of the second author
% \fnmark[2]

% Email id of the second author
\ead{yangfeng@nwpu.edu.cn}

% URL of the second author
\ead[url]{}

% Credit authorship
\credit{Data curation, Methodology, Software}

% Address/affiliation
\affiliation[2]{organization={School of Automation, Northwestern Polytechnical University},
            addressline={}, 
            city={Xi'an},
%          citysep={}, % Uncomment if no comma needed between city and postcode
            postcode={}, 
            state={Shanxi},
            country={China}}

% Corresponding author text
\cortext[1]{Corresponding author}

% Footnote text
\fntext[1]{}

% For a title note without a number/mark
%\nonumnote{}

% Here goes the abstract
\begin{abstract}
Autonomous UAV infiltration in dynamic contested environments remains a significant challenge due to the partially observable nature of threats and the conflicting objectives of mission efficiency versus survivability. Traditional Reinforcement Learning (RL) approaches often suffer from myopic decision-making and struggle to balance these trade-offs in real-time. To address these limitations, this paper proposes an Intent-Context Synergy Reinforcement Learning (ICS-RL) framework. The framework introduces two core innovations: (1) An LSTM-based Intent Prediction Module that forecasts the future trajectories of hostile units, transforming the decision paradigm from reactive avoidance to proactive planning via state augmentation; (2) A Context-Analysis Synergy Mechanism that decomposes the mission into hierarchical sub-tasks (safe cruise, stealth planning, and hostile breakthrough). We design a heterogeneous ensemble of Dueling DQN agents, each specialized in a specific tactical context. A dynamic switching controller based on Max-Advantage values seamlessly integrates these agents, allowing the UAV to adaptively select the optimal policy without hard-coded rules. Extensive simulations demonstrate that ICS-RL significantly outperforms baselines (Standard DDQN) and traditional methods (PSO, Game Theory). The proposed method achieves a mission success rate of 88\% and reduces the average exposure frequency to 0.24 per episode, validating its superiority in ensuring robust and stealthy penetration in high-dynamic scenarios. 
\nocite{*}%% Remove this line from your manuscript.
\end{abstract}

% Use if graphical abstract is present
%\begin{graphicalabstract}
%\includegraphics{}
%\end{graphicalabstract}

% Research highlights
\begin{highlights}
\item A novel ICS-RL framework bridges intent prediction and tactical synergy.
\item LSTM-based state augmentation enables proactive collision avoidance.
\item Advantage-switching mechanism coordinates heterogeneous expert agents.
\item ICS-RL outperforms PSO and Game Theory with a 88\% success rate.
\end{highlights}

%\nocite{*}

% Keywords
% Each keyword is seperated by \sep

\maketitle

% Main text
\section{Introduction}\label{}
In recent years, UAV technology has advanced rapidly. With its high maneuverability, wide attack range, and relatively low cost, UAV have found extensive applications in various fields. Particularly in modern intelligent air combat, they play a crucial role. When performing aerial combat missions, UAV often face more complex environments, making their autonomous decision-making capabilities a key area of research.

Stealth reconnaissance by UAV is a critical task in military warfare. It requires UAV to reach their destination while avoiding detection by the enemy. During this process, scenarios such as obstacle avoidance\cite{Wu2021}, evasion\cite{Jin2024}, and breakthrough\cite{Liu2023} may arise, posing significant challenges to the UAV' maneuvering decision-making capabilities\cite{Wang2024,Liu2021,Xu2025}. To enhance the autonomous decision-making abilities of UAV, extensive research has been conducted, proposing methods such as game theory\cite{Ramirez2018}, optimization algorithms\cite{Duan2015}, and machine learning\cite{Yang2019}.

In game theory research methods, the primary focus is on mathematically modeling maneuvering decision strategies and the movement patterns of UAV. By calculating matrix functions, the optimal strategy is selected\cite{Li2022}. Park et al.\cite{Park2016} proposed an automatic maneuver generation algorithm based on differential game theory for within-visual-range air-to-air combat. This algorithm uses a hierarchical decision structure and score function matrix calculations to generate optimal maneuvers for dynamic and challenging combat environments. Vidal et al.\cite{Vidal2002} studied the theory, implementation, and experimental evaluation of pursuit-evasion games within a probabilistic game theory framework. They proposed two computationally feasible greedy pursuit strategies: local-max and global-max. By implementing these strategies on real UAV and UGVs, they demonstrated their performance under various conditions. However, this approach requires detailed and idealized mathematical expressions of the environment and pursuit strategies in advance. The real-world air combat environment is difficult to deconstruct, and while game theory provides optimal solutions in theory, establishing precise models and the complexity of calculations pose significant challenges when dealing with complex UAV decision-making problems.

Metaheuristic algorithms are a class of computational methods used to solve optimization problems. In the context of UAV decision-making, key parameters such as position, speed, flight path, task allocation, and energy consumption are transformed into decision variables within the optimization problem. An objective function is then established to seek the optimal solution\cite{Duan2023,Najm2019}. Yan et al.\cite{Yan2024} proposed an improved Genetic Algorithm (GA) that integrates multiple constraints to address the task allocation and path planning problems for UAV attacking multiple targets. Shao et al. \cite{Shao2020}introduced an effective path planning method based on a comprehensively improved Particle Swarm Optimization (PSO) algorithm. This method considers flight constraints in complex environments, including terrain avoidance and threat evasion. However, these algorithms typically rely on prior information from a global map for path planning decisions. In real UAV missions, obstacles and enemy information are more random, and these algorithms are prone to getting trapped in local optima.

Nowadays, there is a growing demand for UAV to be more autonomous and intelligent. Reinforcement Learning (RL), an effective machine learning method, offers new solutions for autonomous decision-making and behavior optimization in UAV. RL is a method where an agent continuously interacts with the environment to learn the optimal strategy\cite{Kaelbling1996}. In RL, a UAV (agent) uses various onboard sensors to obtain the environmental state\cite{Liu2019}, executes actions based on the state, and receives feedback (reward) from the environment to learn how to maximize cumulative rewards over the long term\cite{Watkins1992}. This learning approach has inherent advantages in the UAV field, as UAV need to continuously interact with the environment and make adaptive decisions based on environmental changes\cite{Wang2020}.

Sonny et al.\cite{Sonny2023} proposed a Q-learning-based algorithm that can handle both static obstacles and adapt to dynamically changing environments, enabling effective avoidance of dynamic obstacles. They introduced a shortest-distance-first strategy to reduce the distance UAV need to travel to reach their targets. Comparative results showed that this method demonstrated improved performance in learning and path length compared to state-of-the-art path planning methods such as A*, Dijkstra, and SARSA algorithms.

To further enhance UAV decision-making capabilities, researchers have applied methods combining neural networks with reinforcement learning, known as Deep Reinforcement Learning (DRL). In a system proposed by Puente-Castro et al.\cite{Puente-Castro2024}, artificial neural networks (ANN) were used for self-adjustment, learning from both errors and successes to optimize paths. However, the proposed method was not applied to air combat scenarios, overlooking the complexity of stealth reconnaissance missions in real-world air combat situations.

Some researchers (such as Hu et al.\cite{Hu2021}; Yang et al.\cite{Yang2020}; Wang et al.\cite{Wang2022}) have developed intelligent maneuver decision models based on Deep Reinforcement Learning (DRL) and constructed corresponding maneuver libraries, providing UAV with diverse maneuver options. These models utilize the Deep Q-Network (DQN) algorithm, which discretizes maneuver actions to help UAV obtain effective strategies in both close-range and long-range combat, either to defeat opponents or evade attacks. While these methods provide a foundation for air combat models, they make decisions based solely on the current behavior of the enemy, without being able to infer the enemy's intentions and make proactive strategies.

To address the aforementioned issues, this paper proposes the Intent-Context Synergy Reinforcement Learning (ICS-RL) algorithm. To solve the problem of UAV decision-making under different situations within the same task, a Dueling DQN (DDQN) algorithm based on scenario analysis is proposed. This method decomposes different scenarios of the task and trains them separately to derive the optimal strategy. On this basis, the intent analysis method predicts the enemy's possible next actions by analyzing their behavior patterns and historical data, and formulates corresponding evasion strategies, thereby improving the accuracy and effectiveness of decision-making.

The main contributions of this paper are as follows:
\begin{itemize}
    \item A Proactive Intent-Analysis Decision Paradigm: To overcome the myopic limitations of traditional reactive agents, we propose an LSTM-based Intent Prediction Module. By explicitly encoding the enemy's historical trajectory into a future intent representation and augmenting the state space, we transform the UAV's evasion strategy from passive response to proactive planning. This allows the agent to anticipate threats and initiate maneuvering before entering the radar detection range.
    \item A Context-Analysis Synergy Mechanism via Advantage Switching: Addressing the conflicting objectives of mission efficiency and survivability, we establish a hierarchical taxonomy of tactical scenarios (Safe Cruise, Pre-emptive Stealth, Hostile Breakthrough). We design a heterogeneous ensemble of Dueling DRQN agents, each specialized in a specific context. A novel run-time synergy mechanism based on Max-Advantage values is introduced to dynamically delegate control authority, ensuring optimal adaptability without hard-coded rules.
    \item  Superior Performance and Stealth Capability: We validate the proposed ICS-RL framework in a high-fidelity dynamic simulation. Comparative experiments demonstrate that our approach outperforms not only standard DRL baselines but also traditional heuristics (PSO) and game-theoretic methods. Specifically, ICS-RL achieves a 88\% mission success rate and reduces the average exposure frequency to 0.24 per episode, proving its robustness in complex contested environments.
\end{itemize}

\section{Problem Description}
\begin{figure*}[!t]
\centering
\includegraphics[width=.9\textwidth]{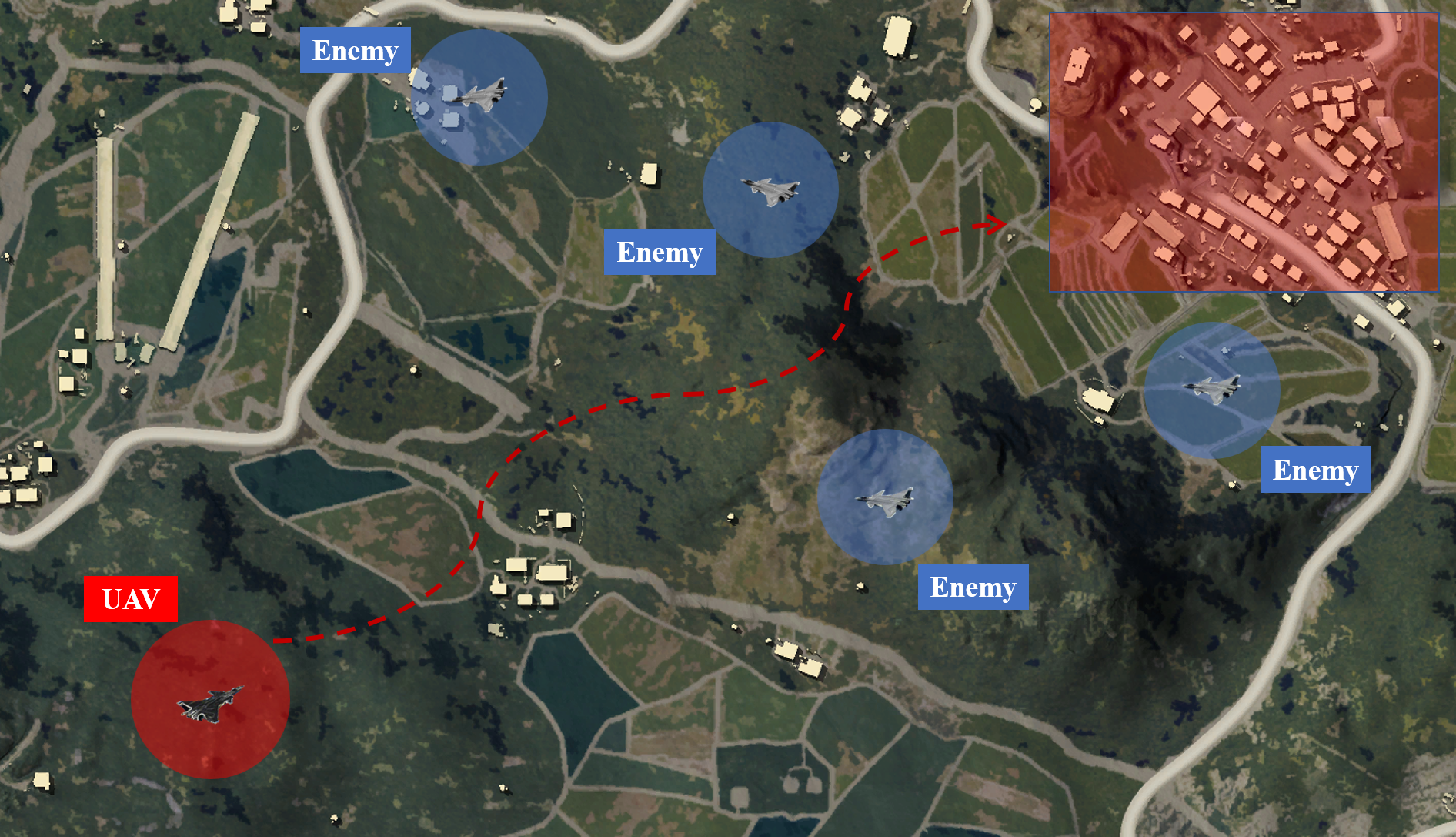}
\caption{Mission Scenario Diagram.}
\label{fig_1}
\end{figure*}

In a battlefield environment, UAV are required to conduct reconnaissance on an enemy target area, necessitating a stealth mission. As shown in Fig.\ref{fig_1}, the red UAV represents our unit, with the red shaded area indicating the detection range of the UAV's sensors, which do not have strike capabilities. Our mission is to reach the target area (the irregular area in the figure) as quickly as possible while avoiding the enemy. The blue UAV represents the enemy unit, with the shaded area indicating its detection range. The enemy can strike if the friendly UAV is detected.

We define \(P_s\) and \(P_e\) as the starting and target positions in the task space, respectively. The flight path of the UAV is composed of N waypoints connecting \(P_s\) and \(P_e\), which can be represented as follows:
\begin{equation}
\label{deqn_ex1}
Tr\{P_s, P_1, P_2, \ldots, P_{i}, P_e\}
\end{equation}

Each waypoint \(P_i\) represents a specific position that the UAV must pass through on its way from the starting position \(P_s\) to the target position \(P_e\). This sequence of waypoints forms the complete flight path for the UAV's mission.

By establishing the following evaluation function, we can measure the "good" or "bad" of UAV actions:

\begin{equation}
\label{deqn_ex1}
F = \alpha_1 C_1 + \alpha_2 C_2 + \alpha_3 C_3 + \cdots + \alpha_M C_M
\end{equation}

where \(C_M\) represents different cost constraints, and \(\alpha_M\) represents the correlation coefficients of the cost constraints \(C_M\). Considering the characteristics of UAV and the safety of the flight process, the evaluation function of the trajectory \(Tr\) can be defined by costs such as flight path length, flight altitude, threats, and physical limitations\cite{Huang2023}. In this paper, the evaluation function is described by selecting the path length cost and threat cost based on specific application scenarios.

\subsection{Path length cost}
The flight distance of the UAV is expected to be as short as possible to demonstrate the efficiency of its decision-making. This cost can be calculated by summing the distances between multiple path nodes. The length of the path nodes is obtained from the Euclidean distance between adjacent points. Therefore, the path length calculation cost of the trajectory \(Tr\) is:

\begin{align}
\label{deqn_ex1}
C_1 &= \sum_{i=0}^{N} \| P_i - P_{i+1} \| 
\end{align}

where \(\| P_i - P_{i+1} \|\) represents the distance the UAV flies from \(P_i\) to \(P_{i+1}\). 

\subsection{Threats cost}
We use the set \( T = \{T_1, T_2, T_3, \ldots, T_k\} \) to define \( k \) threats.

\begin{equation}
\label{deqn_ex1}
\begin{cases} 
C_2 = \sum_{i=0}^{N} \sum_{k=1}^{K} T_k (P_i P_{i+1}) \\
T_k (P_i P_{i+1}) = \\
\begin{cases} 
0, & \|P_i - \text{pos}_k\| > SF \\
SF - \|P_i - \text{pos}_k\|, & 0 < \|P_i - \text{pos}_k\| < SF \\
L, & \|P_i - \text{pos}_k\| = 0 
\end{cases}
\end{cases}
\end{equation}

where \( k \) is the number of threats, \( SF \) is the safety distance determined by factors such as the flight environment and positioning accuracy. \( \text{pos}_k \) is the center of the \( k \)-th threat source, and \( \|P_i - \text{pos}_k\| \) is the distance from the center of the \( k \)-th threat source to the UAV. \( L \) is a penalty constant used to penalize routes that enter dangerous areas.

\section{Theoretical foundation}
\subsection{UAV kinematics model}
\begin{figure}
	\centering
	\includegraphics[width=.9\columnwidth]{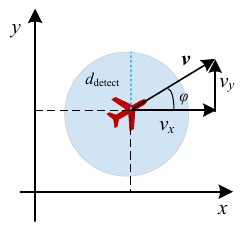}
	\caption{UAV Kinematics model.}
	\label{uavmodel}
\end{figure}
In the research on maneuver decision-making for UAV infiltration missions, the following scenario is constructed. It is assumed that the UAV flies at a fixed altitude and executes coordinated turns with inertia. Fig.\ref{uavmodel} illustrates the two-dimensional kinematic model of the UAV, where the light blue circle represents the UAV's detection range. Consequently, the kinematic equations of the UAV can be expressed as by the following formula:

\begin{equation}
\begin{cases}
p^t = \begin{bmatrix}
x^{t-1} + v_x^t \cdot \Delta t \\
y^{t-1} + v_y^t \cdot \Delta t
\end{bmatrix} \\
v^t = \begin{bmatrix}
v_x^{t-1} + a_x^t \cdot \Delta t \\
v_y^{t-1} + a_y^t \cdot \Delta t
\end{bmatrix} \\
\varphi^t = \text{atan}\left(v_y^t / v_x^t\right)
\end{cases}
\end{equation}

where, \( p^t \), \( v^t \), \( a^t \) and \( \varphi^t \) respectively represents the position, speed, acceleration and yaw angle of UAV at time t, \( \Delta t \) represents the time interval, and \( a \) represents the acceleration of UAV. \( v_x^t \) and \( v_y^t \) respectively represents speed on the X-axis and Y-axis of UAVs at time t.\( a_x^t \) and \( a_y^t \) respectively represents acceleration on the X-axis and Y-axis of UAVs at time t. Considering the limitations of UAV power system and hardware performance, the maximum speed and acceleration are respectively set as \( v_{\max} \) and \( a_{\max} \) in this paper, that is, UAV should meet the performance constraints as Eq.6:

% 公式 (2)
\begin{equation}
\begin{cases}
|v^t| \leq v_{\max} \\
|a^t| \leq a_{\max}
\end{cases}
\end{equation}

\subsection{RL For UAV}
To address the constrained maneuvering decision-making problem for UAV, we propose a reinforcement learning-based method. First, we briefly outline the fundamental theories of RL employed. Then, in Section 4, we provide a detailed description of the proposed reinforcement learning-based decision-making algorithm framework, including the definitions of states, actions, and rewards.

The method for learning the optimal strategy involves Reinforcement Learning (RL)\cite{Puterman1990}. In RL, a UAV (agent) uses various onboard sensors to obtain the environmental state\cite{Sutton1988}, executes actions based on the state, and receives feedback (reward) from the environment. This process allows the UAV to learn how to maximize cumulative rewards over the long term\cite{Liu2015}. This learning approach has inherent advantages in the UAV field, as UAV need to continuously interact with the environment and make adaptive decisions based on environmental changes.

Regardless of the form of reinforcement learning algorithms, their fundamental theoretical framework can be explained using the Markov Decision Process (MDP). The MDP process can be represented as the following quintuple:

\begin{equation}
\label{deqn_ex1}
(S, A, R, TF, \gamma)
\end{equation}

where:
\( S \) represents the set of states in which the agent operates;
\( A \) represents the set of actions the agent can execute;
\( R \) represents the reward provided by the environment, i.e., the return or reward signal the agent receives after interacting with the environment;
\( TF \) is the state transition function, representing the probability distribution of transitioning to the next state \( s_{t+1} \) after the agent executes action \( a_t \) in state \( s_t \);
\( \gamma \) is the discount factor, typically ranging from 0 to 1, which reduces the impact of future rewards and emphasizes the importance of current rewards\cite{Sutton1988}.

\begin{figure}
\centering
\includegraphics[width=.9\columnwidth]{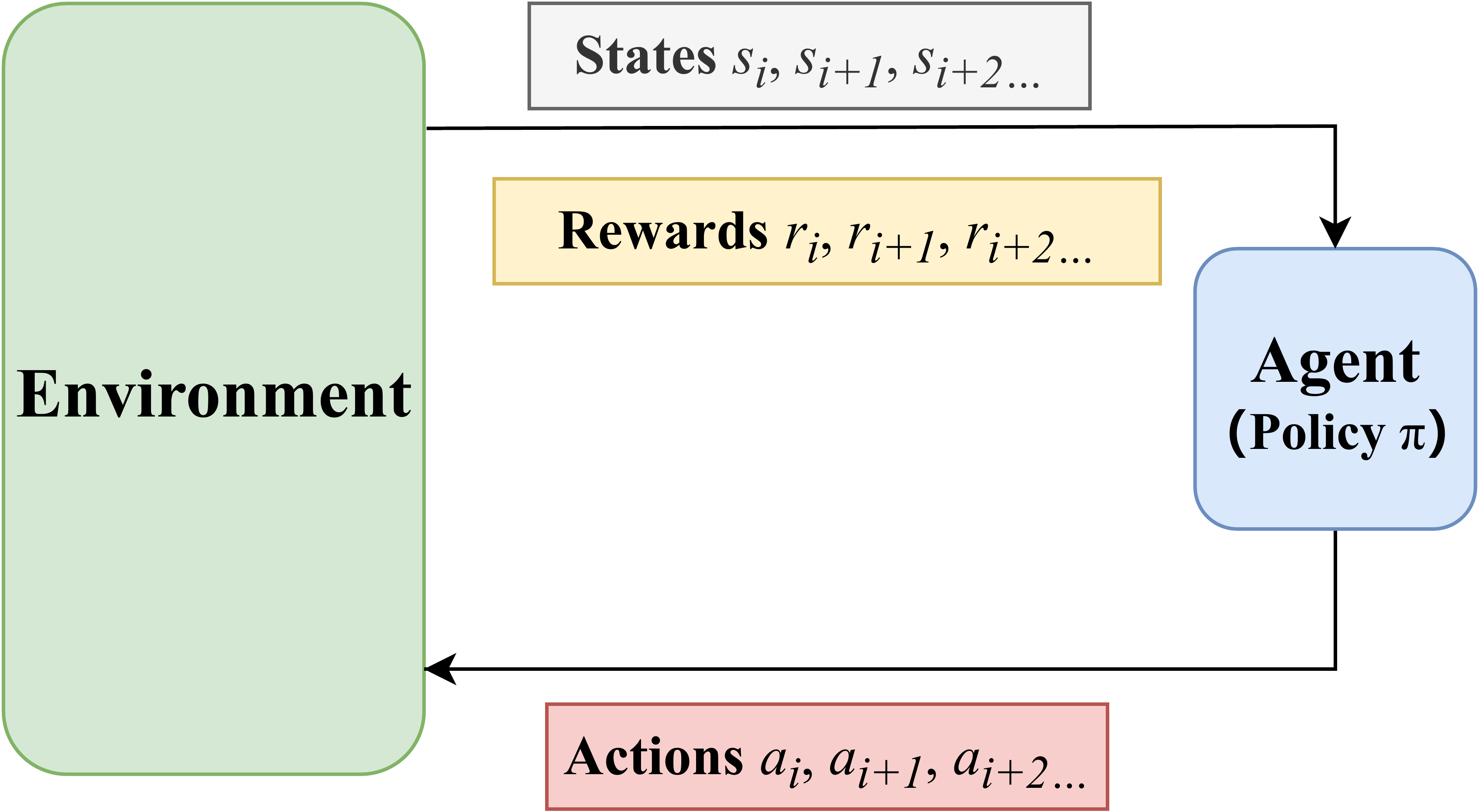}
\caption{Fundamental framework of reinforcement learning.}
\label{fig_2}
\end{figure}

Through repeated interactions, the agent can continuously acquire new states and receive rewards to learn and update its model\cite{Liu2015}, as shown in Fig.\ref{fig_2}.

From the Markov Decision Process, we can deduce that the agent aims to obtain higher rewards through interactions with the environment. Therefore, the goal of reinforcement learning is to construct a policy function that interacts with the environment to maximize the total return within an episode. For the policy function, the value of the state-action pair (s, a) and the state s is defined as follows:

\begin{equation}
\label{deqn_ex1}
\begin{array}{rcl} % 补全花括号{}，列格式rcl正常生效
Q^\pi(s,a) & = & \mathbb{E}\left[\left.R_t\right|s_t=s,a_t=a,\pi\right], \\ % 删除多余的\mathrm
V^\pi(s)   & = & \mathbb{E}_{a\sim\pi(s)}\left[Q^\pi(s,a)\right]
\end{array}
\end{equation}

The state-action value function (commonly referred to as the Q-function) can be recursively computed using dynamic programming. It can be expressed in the following form:

\begin{equation}
\label{deqn_ex1}
Q^{\pi}(s,a)=\mathbb{E}_{s^{\prime}}\left[r+\gamma\mathbb{E}_{a^{\prime}\sim\pi(s^{\prime})}\left[Q^{\pi}(s^{\prime},a^{\prime})\right]\mid s,a,\pi\right]
\end{equation}

Define the optimal \( Q^*(s, a) = \max_{\pi} Q^{\pi}(s, a) \). In a deterministic policy, \( a = \arg\max_{a' \in A} Q^*(s, a') \), we can derive \( V^*(s) = \max_a Q^*(s, a) \). Consequently, the optimal Q-function satisfies the Bellman equation:

\begin{equation}
\label{deqn_ex1}
Q^*(s, a) = \mathbb{E}_{s'} [r + \gamma \max_{a'} Q^*(s', a') | s, a]
\end{equation}

In this research, we focus on a specific quantity, the advantage function, which is related to the Q-function as follows:

\begin{equation}
\label{deqn_ex1}
A^{\pi}(s, a) = Q^{\pi}(s, a) - V^{\pi}(s)
\end{equation}

It is important to note that \( \mathbb{E}_{a \sim \pi(s)} [A^{\pi}(s, a)] = 0 \). This indicates that the value function \( V \) measures the goodness of a particular state \( s \), while the Q-function measures the value of selecting a specific action in that state. The advantage function is obtained by subtracting the value of the state from the Q-function, providing a measure of the importance of each action.

\section{Method}
\begin{figure*}[!t]
\centering
\includegraphics[width=.9\textwidth]{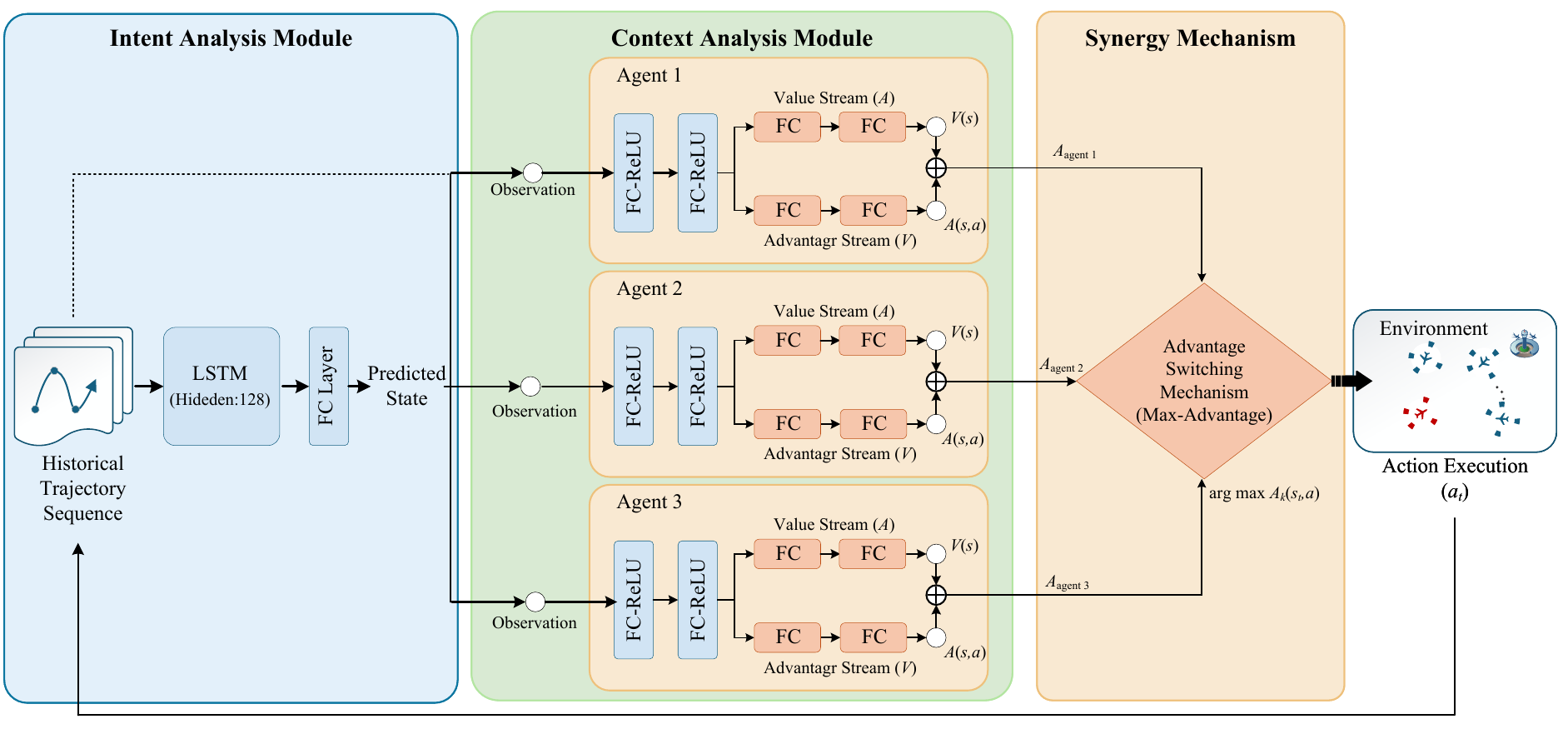}
\caption{The structure diagram of air combat decision model.}
\label{framework}
\end{figure*}

To enhance the combat performance of UAV, a new air combat decision-making model has been designed. It consists of three main components: the Main Environment, the Context Analysis Framework, and the Intent Analysis Framework, as illustrated in the Fig.\ref{framework}.

Main Environment: A DRL (Deep Reinforcement Learning) air combat environment is established for simulation and model training. Based on the UAV task planning problem outlined in Section 2, the action space, state space, and reward function are designed to enable the UAV to complete basic tasks.

Intent Analysis Framework (IA): The purpose of IA is to utilize a prediction module to identify the intentions of enemy units, allowing the friendly UAV to make targeted decisions.

Context Analysis Framework: This mechanism is proposed to train different scenarios separately, improving the selectivity of actions and deriving the optimal strategy, thereby enabling the UAV to better accomplish combat missions.

By integrating these components, the model aims to improve the UAV's decision-making capabilities and overall effectiveness in air combat scenarios.

\subsection{MDP Formulation}
\subsubsection{State Space and Action Space}
State features are the foundation of training. Correct and reasonable features can improve the convergence speed of training and reduce the dimensionality of parameters. Based on the air combat scenario established in this paper, we analyze and design the following state variable tuple to represent the state space of the UAV:

\begin{equation}
\label{deqn_ex1}
S = [P_U(t), V_U(t), P_G, P_E(t), V_E(t), \Delta d]
\end{equation}

where \(P_U(t)\) represents the position information of the friendly UAV, \(V_U(t)\) represents the velocity information of the friendly UAV, \(P_G\) represents the destination coordinate position information, \(P_E(t)\) represents the position information of the enemy UAV, \(V_E(t)\) represents the velocity information of the enemy UAV, \(\Delta d\) represents the Euclidean distance between the enemy unit and the friendly UAV.

To ensure the realism of the flight trajectory and compliance with the UAV's kinematic constraints, we define the action space based on the vehicle's dynamic control inputs rather than direct position jumps. The control input vector at time step $t$ is defined as $u_t = [a_t, \omega_t]^T$, where $a_t$ represents the linear acceleration (controlling velocity changes) and $\omega_t$ represents the angular velocity (controlling heading changes).

\subsubsection{Reward Space}
The reward function is critical for guiding the UAV to learn efficient policies that balance mission completion with survivability. We design a composite reward function $R_t$ composed of four distinct components: navigation reward, threat penalty, boundary penalty, and terminal failure penalty.

Navigation Reward ($R_{nav}$)
To address the sparsity of rewards in large-scale environments, we introduce a potential-based shaping reward combined with a sparse goal achievement reward. The navigation reward is defined as:
\begin{equation}
    R_{nav} = -\lambda_{dis} (d_t - d_{t-1}) + \mathbb{I}_{goal} \cdot C_{dest}
\end{equation}
where $d_t$ denotes the Euclidean distance between the UAV and the target at time step $t$. The term $-\lambda_{dis} (d_t - d_{t-1})$ incentivizes the agent to reduce the distance to the destination at every step. $\mathbb{I}_{goal}$ is an indicator function that equals 1 when the UAV enters the target radius ($d_t < D_{threshold}$), triggering a large positive reward $C_{dest}$.

Threat Penalty ($R_{threat}$)--
To ensure flight safety, the agent receives a penalty whenever it enters the detection range of an enemy unit. This encourages the UAV to plan paths that circumvent hostile radar coverage:
\begin{equation}
    R_{threat} = \begin{cases}
        C_{enemy}, & \text{if } d_{enemy} < D_{detect} \\
        0, & \text{otherwise}
    \end{cases}
\end{equation}
where $d_{enemy}$ is the distance to the nearest enemy unit, $D_{detect}$ is the radar detection range, and $C_{enemy}$ is a negative penalty constant.

Constraint Penalties ($R_{const}$)--
We impose constraints on the flight area and mission durability. A boundary penalty $R_{bound}$ is applied if the UAV leaves the designated mission airspace $\Omega$, and a failure penalty $R_{fail}$ is triggered if the UAV remains under attack for a duration exceeding a safety threshold $T_{safe}$:
\begin{equation}
    R_{const} = \mathbb{I}_{out} \cdot C_{out} + \mathbb{I}_{fail} \cdot C_{fail}
\end{equation}
where $\mathbb{I}_{out} = 1$ if $(x_t, y_t) \notin \Omega$, and $\mathbb{I}_{fail} = 1$ if the accumulated time under threat exceeds $T_{safe}$.

Total Reward--
The final total reward function for the agent at time step $t$ is the summation of the above components:
\begin{equation}
    R_{total} = R_{nav} + R_{threat} + R_{const}
\end{equation}
Specific agents in our ensemble framework may utilize subsets of this reward function (e.g. the evasion expert agent may prioritize $R_{threat}$ over $R_{nav}$) to specialize in distinct tactical behaviors.

\subsection{Intent Analysis}
\label{sec:intent_prediction}

In high-dynamic aerial confrontations, relying solely on instantaneous observations ($S_t$) often leads to myopic decision-making and delayed responses due to the inherent inertia of UAV maneuvers. To overcome this limitation, we introduce an Intent Prediction Module based on Long Short-Term Memory (LSTM) networks. This module serves as a cognitive engine that transforms the problem from a reactive  MDP into a proactive predictive framework.

\subsubsection{Historical Trajectory Representation}
The intent of a hostile unit is implicitly encoded in its historical movement patterns. We define a sliding observation window of length $L$ to capture these temporal dependencies. At time step $t$, the historical trajectory sequence $H_t$ is constructed as:
\begin{equation}
    H_t = \{ \mathbf{o}^{e}_{t-L+1}, \mathbf{o}^{e}_{t-L+2}, \dots, \mathbf{o}^{e}_{t} \}
\end{equation}
where $\mathbf{o}^{e}_{k}$ represents the observed state vector of the enemy (e.g., relative position, velocity, and heading) at step $k$. This sequence $H_t$ serves as the input to the prediction network.

\subsubsection{Temporal Feature Extraction}
We employ an LSTM network to extract deep temporal features from $H_t$. Unlike standard Recurrent Neural Networks (RNNs), LSTMs mitigate the vanishing gradient problem through a gating mechanism, allowing the model to learn long-term dependencies in flight trajectories. The hidden state transition is governed by:
\begin{equation}
    \mathbf{h}_t, \mathbf{c}_t = \text{LSTM}(H_t; \theta_{pred})
\end{equation}
where $\mathbf{h}_t$ is the hidden state vector containing the abstract intent features, $\mathbf{c}_t$ is the cell state, and $\theta_{pred}$ denotes the trainable parameters of the predictor. 

The network then decodes $\mathbf{h}_t$ through a fully connected (FC) layer to forecast the enemy's state at the next time step:
\begin{equation}
    \hat{\mathbf{s}}^{e}_{t+1} = \sigma(W_{out} \mathbf{h}_t + b_{out})
\end{equation}
where $\hat{\mathbf{s}}^{e}_{t+1}$ is the predicted position and heading of the target.

\subsubsection{Supervised Training Objective}
The Intent Prediction Module is trained via supervised learning in parallel with the RL process (or pre-trained). We utilize a Mean Squared Error (MSE) loss function to minimize the discrepancy between the predicted state and the ground truth state recorded in the experience replay buffer:
\begin{equation}
    \mathcal{L}_{pred}(\theta_{pred}) = \frac{1}{M} \sum_{i=1}^{M} \| \hat{\mathbf{s}}^{e}_{i, t+1} - \mathbf{s}^{e}_{i, t+1} \|^2_2
\end{equation}
where $M$ is the batch size sampled from the replay memory.

\subsection{Context Analysis}
\label{ca}
UAV mission scenarios are not static and unchanging. Traditional reinforcement learning methods typically treat scenarios as uniform knowledge for learning, thereby obtaining solutions. Complex penetration missions typically involve conflicting subgoals, such as minimizing flight time and maximizing survival rate. A single monolithic strategy often struggles to balance these trade-offs, leading to convergence to a suboptimal solution. To address this, the ICS-RL framework adopts a "divide-and-conquer" strategy. In Section 2, we have preliminarily decoupled the UAV infiltration process into different situations. We then propose using different agent proxies to treat various situations as distinct domains of knowledge for learning.

To address this, we decompose the infiltration mission into three hierarchical tactical scenarios—safe cruise, pre-emptive stealth, and hostile breakthrough—and design a specialized agent ensemble $\mathcal{K} = \{ \pi_{nav}, \pi_{main}, \pi_{eva} \}$ to master each specific context.

The specialized behaviors of these agents are induced by distinct configurations of the reward weight vector. Initially, in the \textbf{Safe Cruise Phase (Scenario I)}, where the friendly UAV has not detected any enemy units, the primary objective is kinematic efficiency. The \textbf{Navigation Expert ($\pi_{nav}$)} takes precedence in this context. Utilizing a navigation-dominant reward structure that ignores potential threat penalties, this agent optimizes the trajectory solely based on the destination reward ($R_{nav}$), enabling it to compute the shortest, time-optimal path to the target without unnecessary detours.

As the mission progresses to the \textbf{Pre-emptive Stealth Phase (Scenario II)}, characterized by the detection of enemies prior to entering their radar range, the priority shifts to detection avoidance. This context is handled by the \textbf{Stealth Planning Agent ($\pi_{main}$)}. By forecasting the enemy's future trajectory ($\hat{S}_{t+1}$), this agent employs a balanced global reward to plan a path that maintains a safe separation distance, effectively "skirting" the edge of the dynamic radar coverage to balance path deviation with safety.

Finally, in the \textbf{Hostile Breakthrough Phase (Scenario III)}, where the UAV is locked by multiple enemy units, the situation escalates to a survival crisis. The \textbf{Breakthrough Expert ($\pi_{eva}$)} is deployed to break the lock. Trained with a high penalty weight on continuous exposure, $\pi_{eva}$ learns to exploit kinematic vulnerabilities of enemy units (e.g., maximum turning radii or sensor delays) to perform high-G maneuvers, thereby confusing enemy prediction and creating a window for breakthrough.

To seamlessly transition between these distinct phases without relying on hard-coded rules, we employ an \textbf{Advantage-based Dynamic Switching} mechanism. Unlike traditional ensemble methods that average Q-values, we leverage the decomposed streams of the Dueling architecture. At each time step $t$, the system aggregates the advantage streams from all agents and selects the global optimal action $a^*_t$ that maximizes the advantage across the entire ensemble:
\begin{equation}
    a^*_t = \arg \max_{a \in \mathcal{A}} \left( \max_{k \in \{nav, main, eva\}} A_k(s_t, a) \right)
\end{equation}
This mechanism acts as a "contextual switch." For instance, when the UAV is detected (Scenario III), the Breakthrough Expert ($\pi_{eva}$) naturally generates a significantly higher advantage value for sharp evasive turns compared to the Navigation Agent, thus automatically seizing control authority to execute the most relevant maneuver.

By addressing these different scenarios separately, the decision-making model can optimize the UAV's actions and improve its overall mission success rate.

\subsection{Intent-Context Synergy Framework}

The proposed ICS-RL framework integrates the predictive capability of the intent analysis module with the specialized decision-making of the context-Analysis agents. This synergy transforms the UAV's control from a passive response to a proactive intervention strategy.

\subsubsection{Synergy via State Augmentation}
The core innovation lies in explicitly incorporating the enemy's tactical intent into the decision loop. The predicted future state $\hat{\mathbf{s}}^{e}_{t+1}$ is treated as a latent feature of the threat evolution. We augment the Main Agent's observation space by concatenating the current sensory data $S_t$ with this prediction:
\begin{equation}
    S^{aug}_t = [ S_t, \hat{\mathbf{s}}^{e}_{t+1} ]
\end{equation}
This augmented state $S^{aug}_t$ enables the agent to optimize its policy not just based on where the enemy \textit{is}, but where the enemy \textit{will be}, facilitating proactive maneuvers such as interception or pre-emptive evasion.

\subsubsection{Parallel Value Decomposition}
Upon receiving the state inputs (with $S^{aug}_t$ for the Main Agent and $S_t$ for others), the specialized agents in the ensemble $\mathcal{K} = \{nav, main, eva\}$ process the context in parallel. Utilizing the Dueling architecture, each agent $k \in \mathcal{K}$ independently estimates the scalar State Value $V_k(s)$ and the Advantage vector $A_k(s, \cdot)$ for all available actions:

\begin{equation}
\label{eq:duel_outputs}
\begin{cases}
V_{set} = \{ V_{nav}(s), V_{main}(s^{aug}), V_{eva}(s) \} \\
A_{set} = \{ A_{nav}(s, \cdot), A_{main}(s^{aug}, \cdot), A_{eva}(s, \cdot) \}
\end{cases}
\end{equation}
where $A_k(s, \cdot) \in \mathbb{R}^{|\mathcal{A}|}$ represents the relative superiority of each action under the $k$-th agent's specific reward objective.

\subsubsection{Optimal Action Selection and Update}
To execute the optimal strategy, we employ an evaluation module that aggregates these advantages. The system selects the global optimal action $a^*$ by maximizing the advantage value across both the agent ensemble and the action space:
\begin{equation}
\label{eq:action_select}
    a^*_t = \arg \max_{a \in \mathcal{A}} \left( \max_{k \in \mathcal{K}} A_k(s, a) \right)
\end{equation}
This mechanism acts as a dynamic switch, automatically delegating control to the agent with the highest confidence (advantage) in the current situation. 

During the training phase, the chosen agent $k^*$ (which provided the max advantage) is updated using the standard Bellman equation. The Q-value is reconstructed as $Q_{k^*}(s, a^*) = V_{k^*}(s) + A_{k^*}(s, a^*) - \bar{A}_{k^*}(s, \cdot)$, and the network parameters are optimized by minimizing the TD error against the target Q-value stored in the experience replay pool.

\section{Experiments and Analysis}
\label{sec:experiments}

\subsection{Experimental Setup}
\label{subsec:setup}

To comprehensively evaluate the proposed ICS-RL framework, we designed two distinct simulation environments: a \textit{Static Target Scenario} and a \textit{Dynamic Target Scenario} (where the target location refreshes randomly). We conducted comparative experiments against two baselines: the classic DDQN and the Context-Analysis DDQN (CA-DDQN) (an ablation variant without intent prediction). These experiments aim to verify the superiority of the proposed algorithm in terms of strategic learning efficiency, trajectory safety, and intent recognition capability.

\subsubsection{Simulation Environment}
The simulation is conducted on a high-fidelity platform configured with an Intel Core i7 CPU and NVIDIA GTX 1660S GPU. The battlefield spans a $10 \text{ km} \times 10 \text{ km}$ area, containing one friendly UAV, five hostile UAVs, and a designated target zone. 
\begin{itemize}
    \item \textbf{Friendly UAV:} Cruises at $25 \text{ m/s}$ with a sensor detection range of $2 \text{ km}$. Success is defined as entering the $1 \text{ km}$ effective radius of the target.
    \item \textbf{Enemy UAVs:} Patrol outside the target zone at $15 \text{ m/s}$, with a detection range of $1.5 \text{ km}$ and an interception (attack) range of $0.8 \text{ km}$. They employ a heuristic strategy to intercept the friendly UAV upon detection.
\end{itemize}

\begin{table}[h]
\centering
\caption{Hyperparameter Settings for Comparative Algorithms}
\label{tab:params}
\renewcommand{\arraystretch}{1.2} % 稍微增加行高，让表格不那么挤
\begin{tabular}{lccc} 
    \hline  % 双横线模拟粗线效果
    \textbf{Parameter} & \textbf{DDQN} & \textbf{CA-DDQN} & \textbf{ICS-RL (Ours)} \\
    \hline % 单横线
    Replay Buffer Size & $5 \times 10^5$ & $5 \times 10^5$ & $5 \times 10^5$ \\
    Training Episodes  & 40,000          & 40,000          & 40,000          \\
    Learning Rate      & $1 \times 10^{-5}$ & $1 \times 10^{-5}$ & $1 \times 10^{-5}$ \\
    Discount Factor ($\gamma$) & 0.99    & 0.99            & 0.99            \\
    Batch Size         & 512             & 512             & 512             \\
    \hline  % 双横线模拟粗线效果
\end{tabular}
\end{table}

\subsubsection{Hyperparameters}
To ensure a fair comparison, all algorithms share identical training parameters, as detailed in Table \ref{tab:params}.

\subsection{Performance Analysis and Ablation Studies}
\label{subsec:ablation}

To rigorously validate the contribution of each module, we define the \textbf{CA-DDQN} as the ICS-RL framework with the Intent Prediction Module removed. It relies solely on the Context-Analysis Synergy mechanism using the three specialized agents defined in Section \ref{ca}: $\pi_{nav}$ (Scenario 1), $\pi_{main}$ (Scenario 2), and $\pi_{eva}$ (Scenario 3).

\subsubsection{Learning Efficiency and Convergence}
\begin{figure*}
\centering
\includegraphics[width=.9\textwidth]{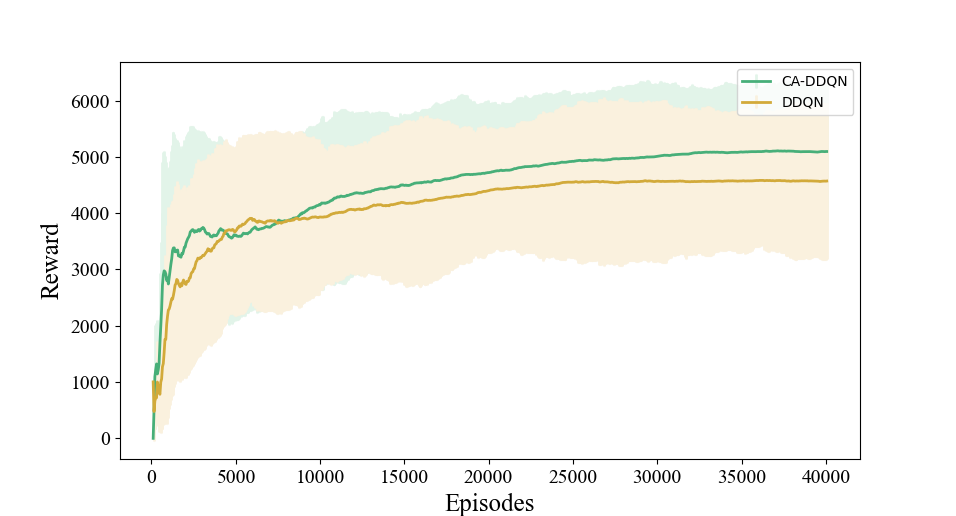}
\caption{Comparison of cumulative reward curves between CA-DDQN and Standard DDQN.}
\label{fig:ca_vs_ddqn}
\end{figure*}

We first compare the learning curves of CA-DDQN and the vanilla DDQN in Fig. \ref{fig:ca_vs_ddqn}. 
While both methods start with low rewards due to initial exploration penalties, CA-DDQN demonstrates a significantly faster convergence rate and achieves a higher asymptotic reward. This improvement is attributed to the \textit{Divide-and-Conquer} strategy: by decomposing the complex mission into simpler sub-tasks (navigation, avoidance, breakthrough), the specialized agents can learn efficient policies much faster than a single monolithic DDQN agent.

\begin{figure*}
\centering
\includegraphics[width=.9\textwidth]{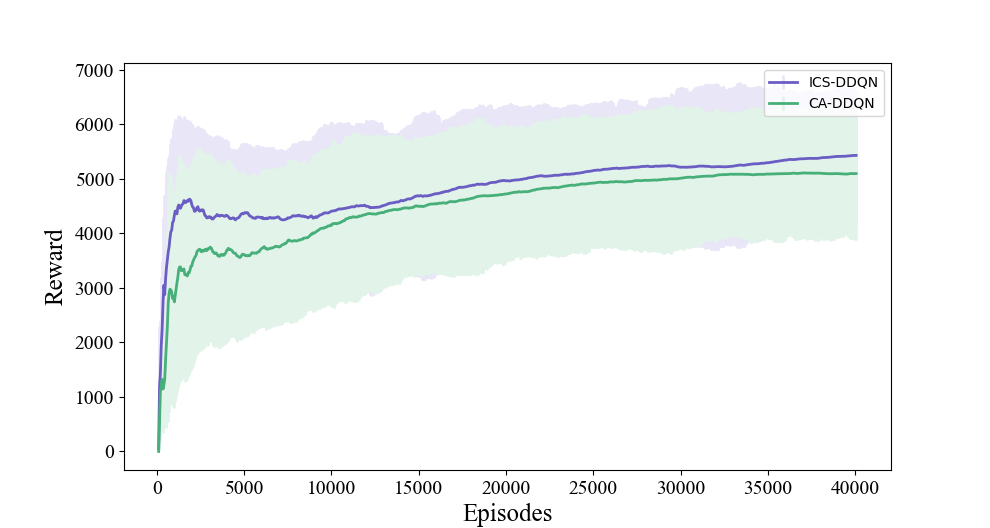}
\caption{Comparison of cumulative reward curves between ICS-DDQN and CA-DDQN.}
\label{fig:ics_vs_ca}
\end{figure*}

Subsequently, we introduce the full ICS-RL (ICS-DDQN) model and compare it against CA-DDQN in Fig. \ref{fig:ics_vs_ca}. The results reveal that ICS-RL not only converges faster (red curve) but also exhibits smaller variance (narrower confidence interval). This stability stems from the \textit{Intent Prediction Module}, which allows the agent to anticipate enemy movements and take pre-emptive actions, thereby reducing the frequency of "surprise" encounters that typically cause large reward penalties and training instability.

\subsubsection{Analysis of Agent Synergy}
\begin{figure}
\centering
 \includegraphics[width=.9\columnwidth]{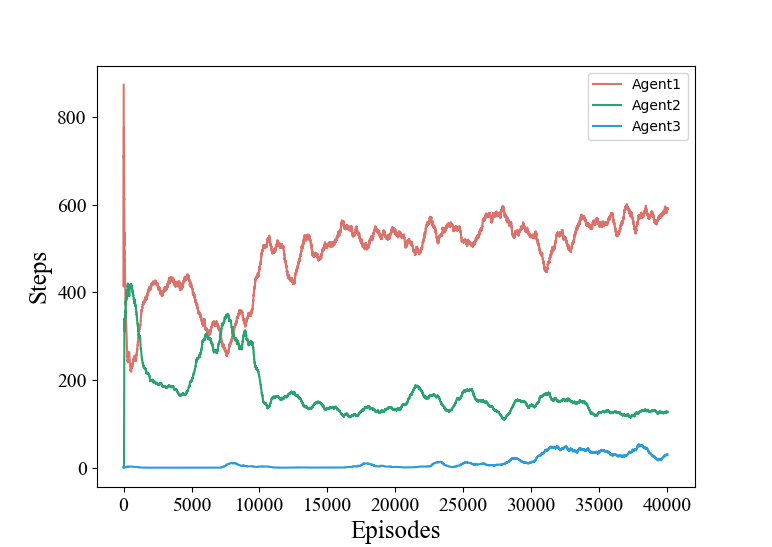}
\caption{Frequency of decision adoption for each specialized agent during training.}
\label{fig:agent_usage}
\end{figure}

To visualize the internal synergy mechanism, Fig. \ref{fig:agent_usage} records the activation frequency of each agent during training.
\begin{itemize}
    \item \textbf{Agent 1 ($\pi_{nav}$):} Dominates the decision-making process, as the majority of the mission time involves cruising in safe zones.
    \item \textbf{Agent 2 ($\pi_{main}$):} Is the second most active, handling the critical phase of detection avoidance.
    \item \textbf{Agent 3 ($\pi_{eva}$):} Is activated least frequently. This is the desired outcome, as it indicates that the system effectively avoids entering the dangerous "locked-on" state (Scenario 3) by successfully handling threats in the earlier "detection" phase (Scenario 2).
\end{itemize}

\begin{figure}
\centering
 \includegraphics[width=.9\columnwidth]{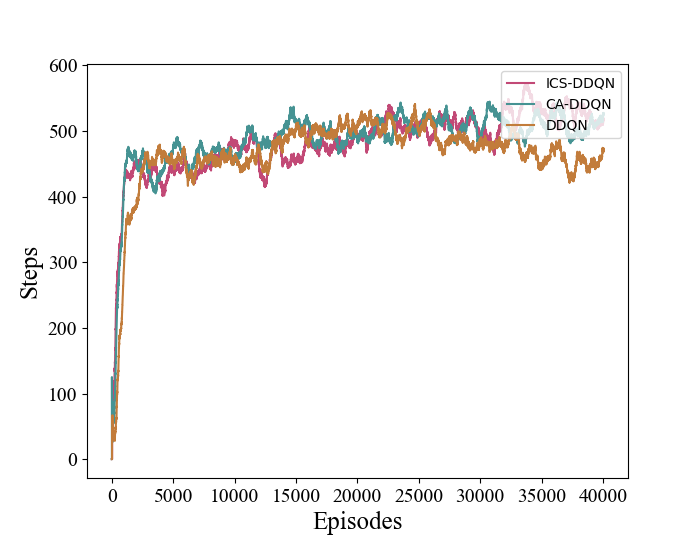}
\caption{Comparison of positive reward steps, indicating mission completion speed.}
\label{fig:positive_steps}
\end{figure}

\subsubsection{Mission Efficiency Metrics}
We further evaluate mission efficiency using the "Number of Positive Reward Steps" metric (Fig. \ref{fig:positive_steps}). Since positive rewards are sparse and only awarded near the target, a higher frequency of positive steps earlier in training indicates that ICS-RL discovers the path to the destination much faster than DDQN. This confirms that the proposed method significantly reduces the time-to-target.

\subsection{Comparative Testing Results}
To comprehensively evaluate the adaptability of the proposed ICS-RL framework against distinct algorithmic paradigms, we expanded our baseline comparison to include two traditional non-learning methods:
\begin{itemize}
    \item \textbf{Particle Swarm Optimization (PSO):} A heuristic optimization algorithm that plans trajectories by iteratively improving candidate solutions based on a fitness function.\cite{Khargharia_Ouali_Shakya_Ahmad_2026}
    \item \textbf{Game Theory (GT):} A decision-making approach that calculates the Nash Equilibrium strategy assuming rational opponents, often used in adversarial discrete scenarios\cite{Zhou_Li_Sheng_Qi_Cong_2024}.
\end{itemize}

\begin{figure}
\centering
\includegraphics[width=.9\columnwidth]{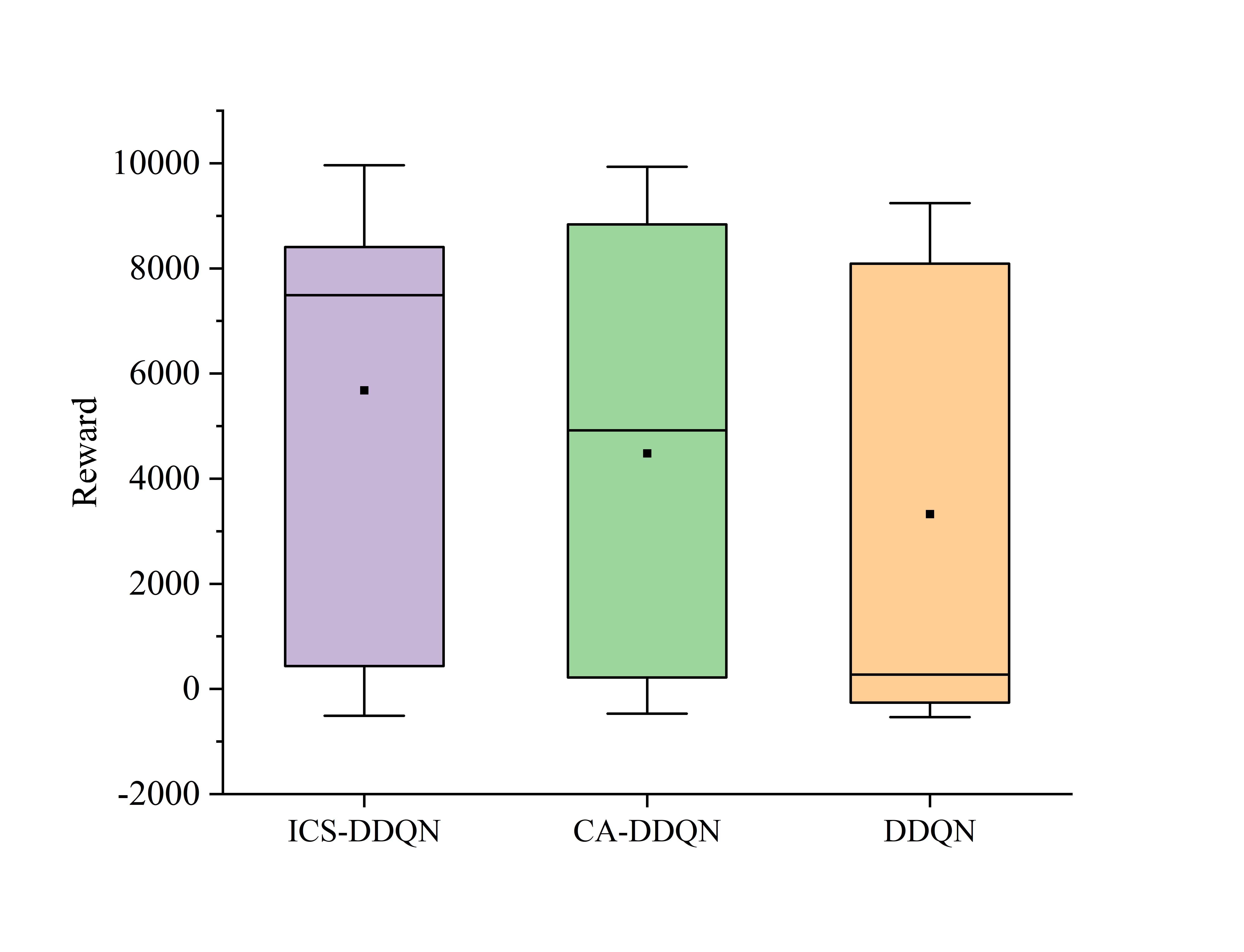}
\caption{Statistical distribution of total rewards over 50 test episodes.}
\label{fig:test_rewards}
\end{figure}

\subsubsection{Comprehensive Performance Assessment}
We conducted 50 independent Monte Carlo simulation runs for each method. The comprehensive performance metrics, including Success Rate (SR), Average Exposure Count (AEC, indicating stealth capability), and Prediction Accuracy(PA), are summarized in Table \ref{tab:comprehensive_baselines}.

\begin{table}[width=.9\linewidth,cols=4,pos=h]
    \centering
    \caption{Performance Comparison with Traditional and DRL Algorithms}
    \label{tab:comprehensive_baselines}
    \setlength{\tabcolsep}{3pt} % Adjust column spacing
    \begin{tabular*}{\tblwidth}{@{} LLLL@{} }
        \toprule
        \textbf{Method } & \textbf{SR} & \textbf{AEC} & \textbf{PA} \\
        \midrule      
        PSO  & 69.0\% & 1.87 & N/A \\
        Game Theory & 77.0\% & 1.41 & N/A \\
         Standard DDQN & 64\% & 1.56 & N/A \\
        CA-DDQN (Ours) & 80\% & 1.15 & N/A \\
        \textbf{ICS-RL (Ours)} & \textbf{88\%} & \textbf{0.24} & \textbf{80.2\%} \\
        \bottomrule
    \end{tabular*}
\end{table}

The statistical results are presented in Fig. \ref{fig:test_rewards}.
The box plot clearly shows that ICS-RL achieves the highest median reward and the smallest interquartile range (IQR), indicating robust and consistently high performance. In contrast, the standard DDQN exhibits a wide spread with a much lower median, reflecting its instability in complex dynamic scenarios.

\begin{figure*}
        \center
        \scriptsize
        \begin{tabular}{cccc}
                \includegraphics[width=1.5in]{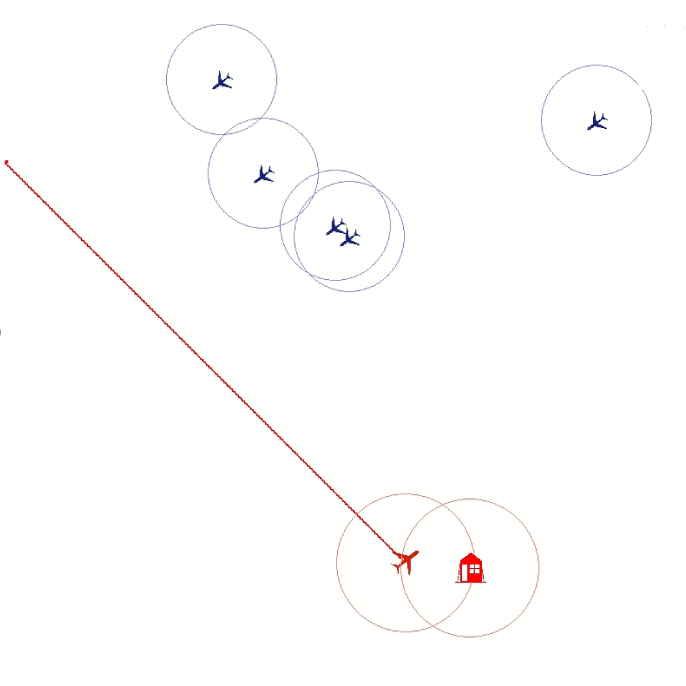} &    \includegraphics[width=1.5in]{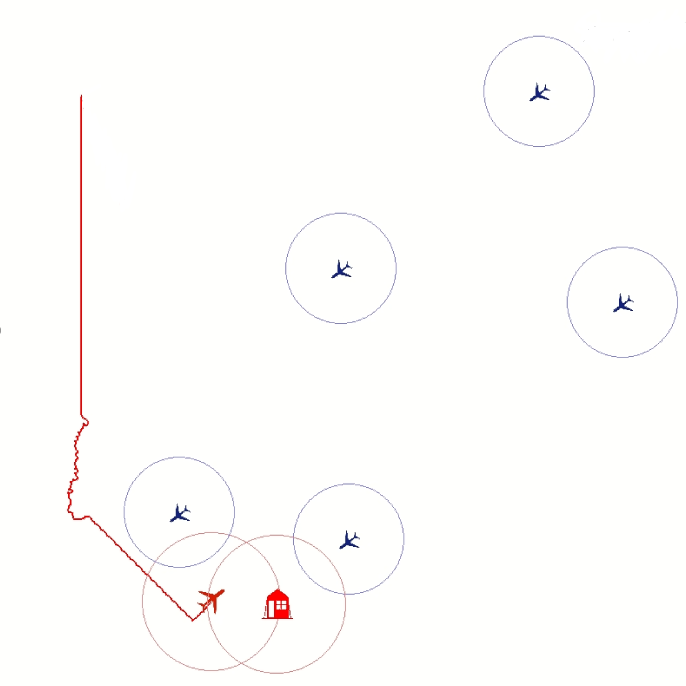}  &    \includegraphics[width=1.5in]{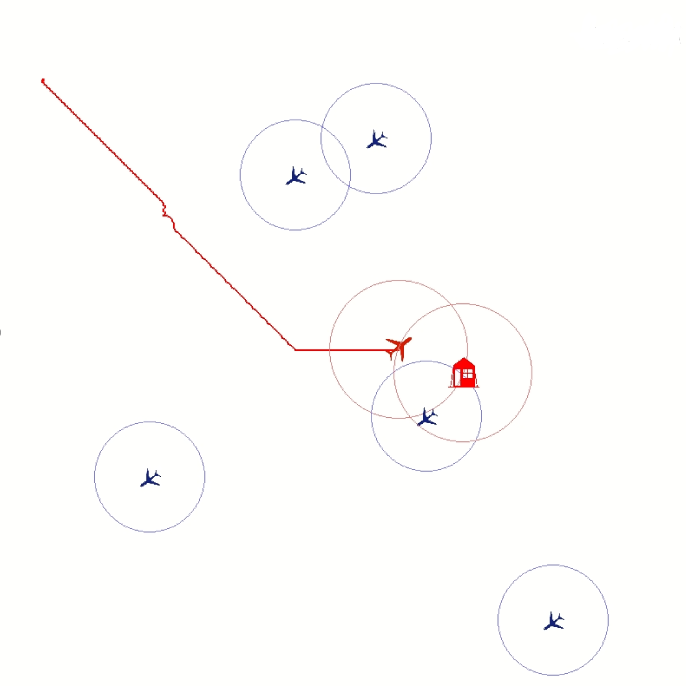}   &    \includegraphics[width=1.5in]{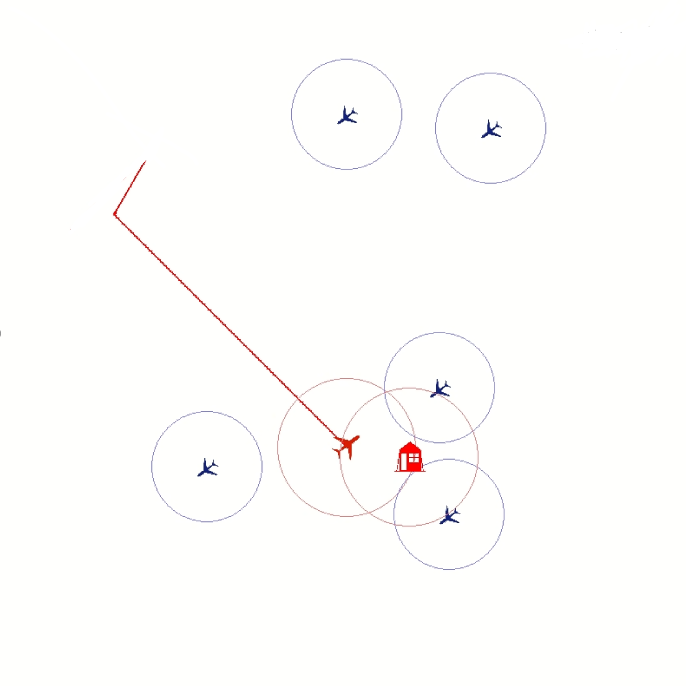}     \\
                (a) & (b)& (c)& (d) \\
                 \includegraphics[width=1.5in]{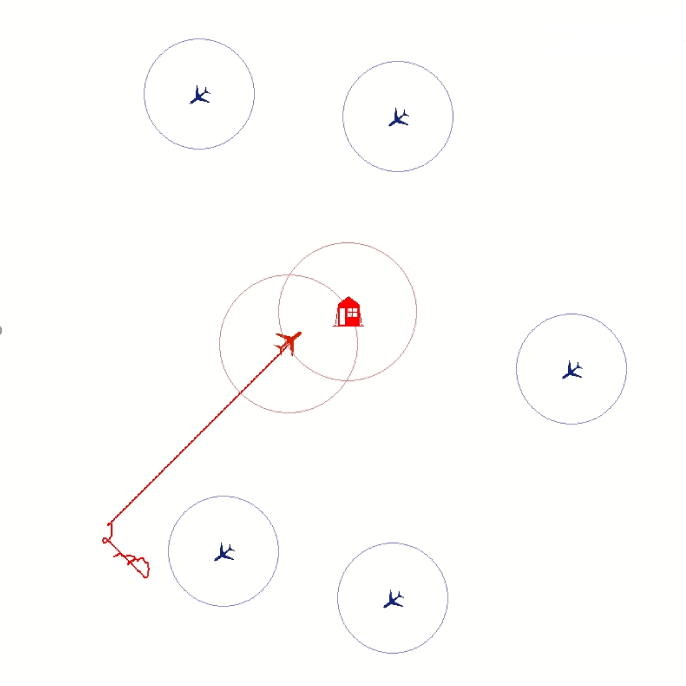} &    \includegraphics[width=1.5in]{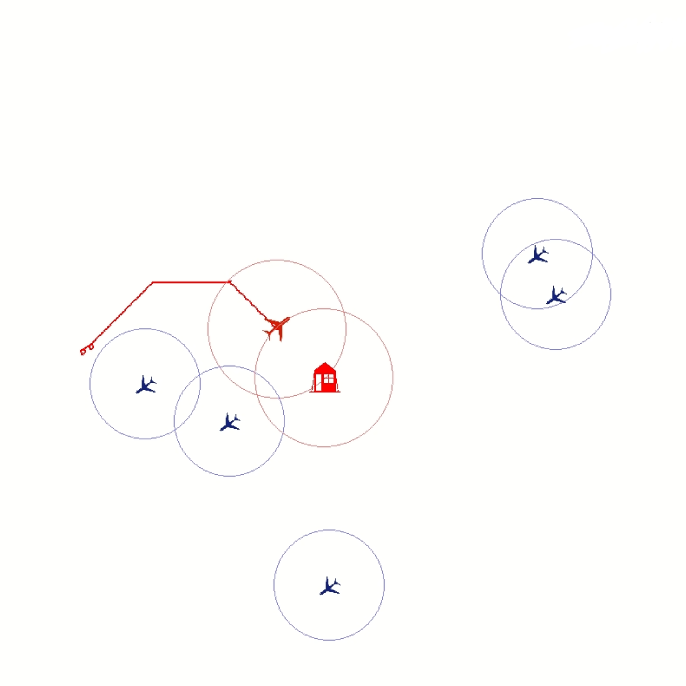}  &    \includegraphics[width=1.5in]{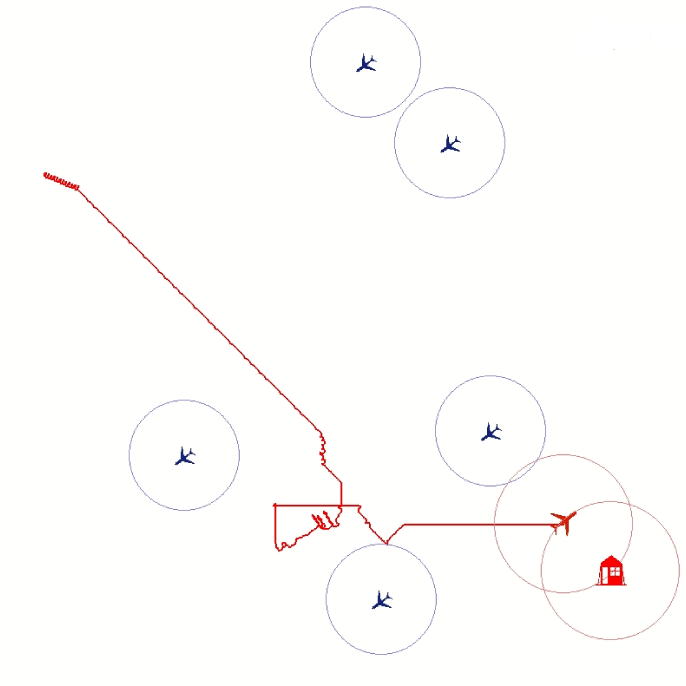}   &    \includegraphics[width=1.5in]{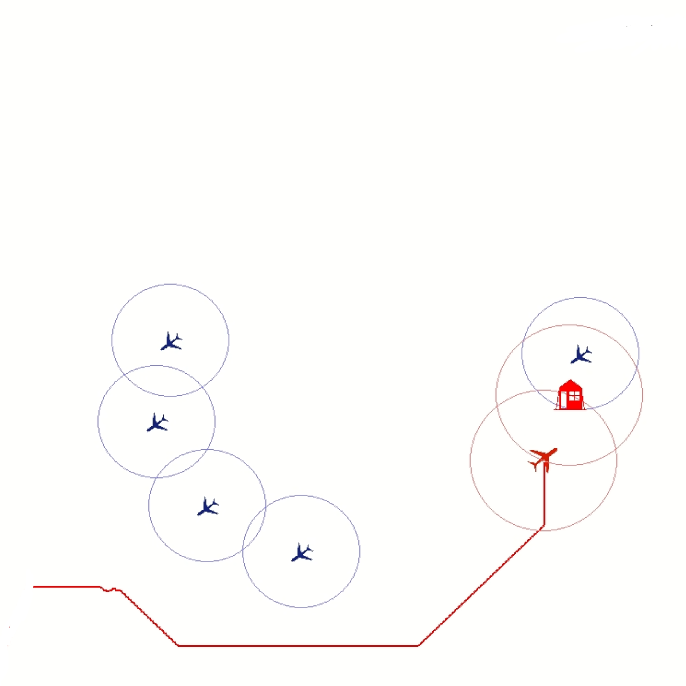} \\
                (e) & (f)& (g)& (h) \\
        \end{tabular}
        \caption{Detail of a trajectory performed by ICS-DDQN}
        \label{figure16}
        \vspace{-0.5em}
\end{figure*}

\subsubsection{Analysis of Traditional vs. Learning Methods}
As observed in Table \ref{tab:comprehensive_baselines}, while traditional methods outperform the vanilla DQN, they still fall short of the proposed ICS-RL framework.

\textbf{PSO Analysis:} The PSO algorithm achieves a success rate of 69.0\%. While it performs better than the standard DQN (64.5\%) due to its global search capability in static snapshots, it struggles in highly dynamic environments. PSO often converges to local optima when the enemy constellation changes rapidly, leading to a relatively high average exposure time of 1.87s.

\textbf{Game Theory Analysis:} The Game Theory approach demonstrates strong performance with a 77.0\% success rate and a lower exposure time of 1.41s. This is because it effectively models the adversarial nature of the problem. However, its reliance on discrete state spaces and the assumption of perfect opponent rationality limits its flexibility against stochastic behaviors. Furthermore, both PSO and GT lack a temporal prediction mechanism (Pred. Accuracy = 0.00\%), forcing them to react to threats only after they manifest.

\textbf{Superiority of ICS-RL:} In contrast, the ICS-RL framework achieves the highest success rate (88\%) and the lowest exposure time ($\sim$0.24s). This superiority stems from the \textit{Intent Prediction Module}, which provides high-fidelity trajectory forecasting (80.2\% accuracy). This predictive capability enables the agent to execute proactive maneuvers that traditional "reactive" or "equilibrium-based" methods cannot anticipate.

\subsection{Behavior Analysis}

We present typical maneuver decision scenarios based on the proposed CA and IA methods in Figure 16. Figure 16(a) represents a relatively simple task scenario, where there are no enemy UAV (blue) near either the initial position of the friendly UAV (red) or the target area to be infiltrated. As a result, the friendly UAV flies directly to the target area along the shortest path. As shown in Figures 16(b) and (c), enemy drones are present near the target area. In (b), an enemy UAV is located on the path of the friendly UAV to the target area. the friendly UAV determines that it can enter the target area before the enemy intercepts it. In (c), the red UAV enters the target area before being detected by the enemy UAV. In the encounter scenarios shown in (d) to (f), enemy units are already present near the initial position of the red UAV and are approaching it. the friendly UAV analyzes the enemy's intentions and, consequently, makes an evasive decision to avoid detection by the enemy. It then identifies a safer path to fly toward the target area. In (g), during flight, the red UAV detects that two enemy drones are forming an encirclement. As a result, it promptly makes a detour decision, waiting for the enemy to depart before continuing its journey and ultimately reaching the target area. In the scenario shown in (h), four enemy drones are approaching in a blocking formation. the friendly UAV detects the enemy's tactical intentions and, before being cornered, identifies a gap in the enemy's blockade. It promptly escapes and successfully flies toward the target area while avoiding detection by one of the defending UAV.

By analyzing the UAV decision-making behavior in various scenarios, it is fully demonstrated that the proposed CA framework, by decoupling different situations, guides the agent to achieve higher rewards and make better decisions. The proposed IA framework helps the agent anticipate the enemy's behavioral intentions and make decisions that avoid penalties. The combination of both methods allows the friendly UAV to complete tasks more efficiently and safely.

\section{Conclusion}
This paper investigates the maneuvering decision-making problem for UAV infiltration reconnaissance in air combat. First, a mathematical expression of the UAV task model is established and transformed into a constrained optimization problem. Within the reinforcement learning framework, air combat data is converted into learnable samples. Finally, to address the issue of UAV being unable to effectively judge enemy intentions and adapt to different situations, we propose an ICS-RL air combat decision-making model. This model is based on intent analysis and scenario analysis methods.

The scenario analysis mechanism introduced in this paper refines the interaction between the agent and the environment, expanding the action strategy space and effectively reducing the likelihood of suboptimal action strategies. The improved intent analysis reinforcement learning method enables the UAV to predict the enemy unit's next move, helping the UAV to make preemptive decisions and effectively avoid danger. This advancement reduces the combat risk for the UAV and the probability of being detected during infiltration missions, providing significant operational advantages.

Simulation results show that agents trained with this algorithm can achieve safer and more effective maneuvering strategies compared to the original algorithm, with a higher success rate in completing missions.

% Numbered list
% Use the style of numbering in square brackets.
% If nothing is used, default style will be taken.
%\begin{enumerate}[a)]
%\item 
%\item 
%\item 
%\end{enumerate}  

% Unnumbered list
%\begin{itemize}
%\item 
%\item 
%\item 
%\end{itemize}  

% Description list
%\begin{description}
%\item[]
%\item[] 
%\item[] 
%\end{description}  

% Uncomment and use as the case may be
%\begin{theorem} 
%\end{theorem}

% Uncomment and use as the case may be
%\begin{lemma} 
%\end{lemma}

%% The Appendices part is started with the command \appendix;
%% appendix sections are then done as normal sections
%% \appendix

% To print the credit authorship contribution details
\printcredits

%% Loading bibliography style file
%\bibliographystyle{model1-num-names}

\bibliographystyle{cas-model2-names}

% Loading bibliography database
\bibliography{cas-refs}

@article{Wu2021,
  author  = {J. Wu and C. Luo and Y. Luo and K. Li},
  title   = {Distributed UAV swarm formation and collision avoidance strategies over fixed and switching topologies},
  journal = {IEEE Trans. Cybern.},
  year    = {2021},
  volume  = {52},
  number  = {10},
  pages   = {10969--10979}
}

@article{Jin2024,
  author  = {W. Jin and X. Tian and B. Shi and B. Zhao and H. Duan and H. Wu},
  title   = {Enhanced UAV Pursuit-Evasion Using Boids Modelling: A Synergistic Integration of Bird Swarm Intelligence and DRL},
  journal = {Comput. Mater. Continua},
  year    = {2024},
  volume  = {80},
  number  = {3}
}

@inproceedings{Liu2023,
  author    = {S. Liu and T. Chen and T. Zhao and S. Liu and C. Ma},
  title     = {Research on cooperative UAV countermeasure strategy based on interception triangle},
  booktitle = {Proceedings of the 2023 4th International Conference on Machine Learning and Computer Application},
  year      = {2023},
  pages     = {1015--1020},
  month     = {Oct}
}

@article{Wang2024,
  author  = {X. Wang and Y. Wang and X. Su and L. Wang and C. Lu and H. Peng and J. Liu},
  title   = {Deep reinforcement learning-based air combat maneuver decision-making: literature review, implementation tutorial and future direction},
  journal = {Artif. Intell. Rev.},
  year    = {2024},
  volume  = {57},
  number  = {1},
  pages   = {1}
}

@article{Liu2021,
  author  = {C. Liu and S. Sun and C. Tao and Y. Shou and B. Xu},
  title   = {Sliding mode control of multi-agent system with application to UAV air combat},
  journal = {Comput. Electr. Eng.},
  year    = {2021},
  volume  = {96},
  pages   = {107491}
}

@article{Xu2025,
  author  = {X. Xu and Y. Wang and X. Guo and K. Huang and X. Zhang},
  title   = {Multi-UAV air combat cooperative game based on virtual opponent and value attention decomposition policy gradient},
  journal = {Expert Syst. Appl.},
  year    = {2025},
  volume  = {267},
  pages   = {126069}
}

@article{Ramirez2018,
  author  = {N. Ramírez López and R. Żbikowski},
  title   = {Effectiveness of autonomous decision making for unmanned combat aerial vehicles in dogfight engagements},
  journal = {J. Guid. Control Dyn.},
  year    = {2018},
  volume  = {41},
  number  = {4},
  pages   = {1021--1024}
}

@article{Duan2015,
  author  = {H. Duan and P. Li and Y. Yu},
  title   = {A predator-prey particle swarm optimization approach to multiple UCAV air combat modeled by dynamic game theory},
  journal = {IEEE/CAA J. Automatica Sinica},
  year    = {2015},
  volume  = {2},
  number  = {1},
  pages   = {11--18}
}

@article{Yang2019,
  author  = {Q. Yang and J. Zhang and G. Shi and J. Hu and Y. Wu},
  title   = {Maneuver decision of UAV in short-range air combat based on deep reinforcement learning},
  journal = {IEEE Access},
  year    = {2019},
  volume  = {8},
  pages   = {363--378}
}

@article{Li2022,
  author  = {S. Y. Li and M. Chen and Y. H. Wang and Q. X. Wu},
  title   = {Air combat decision-making of multiple UCAVs based on constraint strategy games},
  journal = {Def. Technol.},
  year    = {2022},
  volume  = {18},
  number  = {3},
  pages   = {368--383}
}

@article{Park2016,
  author  = {H. Park and B.-Y. Lee and M.-J. Tahk and D.-W. Yoo},
  title   = {Differential Game Based Air Combat Maneuver Generation Using Scoring Function Matrix},
  journal = {Int. J. Aeronaut. Space Sci.},
  year    = {2016},
  volume  = {17},
  number  = {2},
  pages   = {204--213},
  doi     = {10.5139/IJASS.2016.17.2.204}
}

@article{Vidal2002,
  author  = {R. Vidal and O. Shakernia and H. J. Kim and D. H. Shim and S. Sastry},
  title   = {Probabilistic pursuit-evasion games: Theory, implementation, and experimental evaluation},
  journal = {IEEE Trans. Robot. Autom.},
  year    = {2002},
  volume  = {18},
  number  = {5},
  pages   = {662--669},
  doi     = {10.1109/TRA.2002.804040}
}

@article{Duan2023,
  author  = {H. Duan and Y. Lei and J. Xia and Y. Deng and Y. Shi},
  title   = {Autonomous Maneuver Decision for Unmanned Aerial Vehicle via Improved Pigeon-Inspired Optimization},
  journal = {IEEE Trans. Aerosp. Electron. Syst.},
  year    = {2023},
  volume  = {59},
  number  = {3},
  pages   = {3156--3170},
  month   = {June},
  doi     = {10.1109/TAES.2022.3221691}
}

@article{Najm2019,
  author  = {A. A. Najm and I. K. Ibraheem},
  title   = {Nonlinear PID controller design for a 6-DOF UAV quadrotor system},
  journal = {Eng. Sci. Technol. Int. J.},
  year    = {2019},
  volume  = {22},
  number  = {4},
  pages   = {1087--1097},
  doi     = {10.1016/j.jestch.2019.02.005}
}

@article{Yan2024,
  author  = {F. Yan and J. Chu and J. Hu and X. Zhu},
  title   = {Cooperative task allocation with simultaneous arrival and resource constraint for multi-UAV using a genetic algorithm},
  journal = {Expert Syst. Appl.},
  year    = {2024},
  volume  = {245},
  pages   = {123023},
  doi     = {10.1016/j.eswa.2023.123023}
}

@article{Shao2020,
  author  = {S. Shao and Y. Peng and C. He and Y. Du},
  title   = {Efficient path planning for UAV formation via comprehensively improved particle swarm optimization},
  journal = {ISA Trans.},
  year    = {2020},
  volume  = {97},
  pages   = {415--430},
  doi     = {10.1016/j.isatra.2019.08.018}
}

@article{Kaelbling1996,
  author  = {L. P. Kaelbling and M. L. Littman and A. W. Moore},
  title   = {Reinforcement learning: A survey},
  journal = {J. Artif. Intell. Res.},
  year    = {1996},
  volume  = {4},
  pages   = {237--285}
}

@article{Liu2019,
  author  = {X. Liu and Y. Liu and Y. Chen},
  title   = {Reinforcement Learning in Multiple-UAV Networks: Deployment and Movement Design},
  journal = {IEEE Trans. Veh. Technol.},
  year    = {2019},
  volume  = {68},
  number  = {8},
  pages   = {8036--8049},
  month   = {Aug},
  doi     = {10.1109/TVT.2019.2922849}
}

@article{Watkins1992,
  author  = {C. J. C. H. Watkins and P. Dayan},
  title   = {Q-learning},
  journal = {Mach. Learn.},
  year    = {1992},
  volume  = {8},
  pages   = {279--292},
  doi     = {10.1007/BF00992698}
}

@article{Wang2020,
  author  = {C. Wang and J. Wang and J. Wang and X. Zhang},
  title   = {Deep-Reinforcement-Learning-Based Autonomous UAV Navigation With Sparse Rewards},
  journal = {IEEE Internet Things J.},
  year    = {2020},
  volume  = {7},
  number  = {7},
  pages   = {6180--6190},
  month   = {July},
  doi     = {10.1109/JIOT.2020.2973193}
}

@article{Sonny2023,
  author  = {A. Sonny and S. R. Yeduri and L. R. Cenkeramaddi},
  title   = {Q-learning-based unmanned aerial vehicle path planning with dynamic obstacle avoidance},
  journal = {Appl. Soft Comput.},
  year    = {2023},
  volume  = {147},
  pages   = {110773},
  doi     = {10.1016/j.asoc.2023.110773}
}

@article{Puente-Castro2024,
  author  = {Puente-Castro, Alejandro and Rivero, Daniel and Pedrosa, Eurico and Pereira, Artur and Lau, Nuno and Fernandez-Blanco, Enrique},
  title   = {Q-Learning based system for Path Planning with Unmanned Aerial Vehicles swarms in obstacle environments},
  journal = {Expert Syst. Appl.},
  year    = {2024},
  volume  = {235},
  pages   = {121240},
  doi     = {10.1016/j.eswa.2023.121240}
}

@article{Hu2021,
  author  = {D. Hu and R. Yang and J. Zuo and Z. Zhang and J. Wu and Y. Wang},
  title   = {Application of deep reinforcement learning in maneuver planning of beyond-visual-range air combat},
  journal = {IEEE Access},
  year    = {2021},
  volume  = {9},
  pages   = {32282--32297},
  doi     = {10.1109/ACCESS.2021.3060426}
}

@article{Yang2020,
  author  = {Q. Yang and J. Zhang and G. Shi and J. Hu and Y. Wu},
  title   = {Maneuver decision of uav in short-range air combat based on deep reinforcement learning},
  journal = {IEEE Access},
  year    = {2020},
  volume  = {8},
  pages   = {363--378},
  doi     = {10.1109/ACCESS.2019.2961426}
}

@article{Wang2022,
  author  = {X. Wang and H. Peng and J. Liu and X. Dong and X. Zhao and C. Lu},
  title   = {Optimal control based coordinated taxiing path planning and tracking for multiple carrier aircraft on flight deck},
  journal = {Def. Technol.},
  year    = {2022},
  volume  = {18},
  pages   = {238--248},
  doi     = {10.1016/j.dt.2020.11.013}
}

@article{Huang2023,
  author  = {C. Huang and X. Zhou and X. Ran and J. Wang and H. Chen and W. Deng},
  title   = {Adaptive cylinder vector particle swarm optimization with differential evolution for UAV path planning},
  journal = {Eng. Appl. Artif. Intell.},
  year    = {2023},
  volume  = {121},
  pages   = {105942},
  doi     = {10.1016/j.engappai.2023.105942}
}

@article{Puterman1990,
  author  = {M. L. Puterman},
  title   = {Markov decision processes},
  journal = {Handb. Oper. Res. Manag. Sci.},
  year    = {1990},
  volume  = {2},
  pages   = {331--434}
}

@article{Sutton1988,
  author  = {R. S. Sutton},
  title   = {Learning to predict by the methods of temporal differences},
  journal = {Mach. Learn.},
  year    = {1988},
  volume  = {3},
  number  = {1},
  pages   = {9--44}
}

@article{Liu2015,
  author  = {C. Liu and X. Xu and D. Hu},
  title   = {Multiobjective Reinforcement Learning: A Comprehensive Overview},
  journal = {IEEE Trans. Syst. Man Cybern. Syst.},
  year    = {2015},
  volume  = {45},
  number  = {3},
  pages   = {385--398},
  month   = {March},
  doi     = {10.1109/TSMC.2014.2358639}
}

@article{Zhou_Li_Sheng_Qi_Cong_2024, 
title={Optimal control strategies and target selection in multi-pursuer multi-evader differential games}, 
volume={588}, 
ISSN={09252312}, 
DOI={10.1016/j.neucom.2024.127701},  
journal={Neurocomputing}, 
author={Zhou, Yinglu and Li, Yinya and Sheng, Andong and Qi, Guoqing and Cong, Jinliang}, 
year={2024}, 
month=july, 
pages={127701}, 
language={en} }

@article{Khargharia_Ouali_Shakya_Ahmad_2026, 
title={Collision avoidance in UAV swarms: A learning-centric perspective on collaborative intelligence}, volume={663}, ISSN={0925-2312}, DOI={https://doi.org/10.1016/j.neucom.2025.132020}, abstractNote={As UAV swarm deployments become more prevalent in mission critical domains, collision avoidance remains a key challenge in ensuring safety, coordination, and autonomy at scale. This survey investigates the state of the art in learning based collision avoidance strategies enabled through collaborative intelligence in UAV swarms. We introduce a six dimensional taxonomy that classifies approaches across decision making paradigms, swarm coordination models, communication architectures, learning methodologies, execution strategies, and safety assurance mechanisms. The survey places particular emphasis on learning based methodologies, which we categorize into four prominent techniques: reinforcement learning, federated learning, neuro inspired models, and hybrid approaches. For each, we provide a detailed review of training architectures, scalability, robustness, and real-time feasibility. Drawing on peer-reviewed publications (2019 to early 2025), we synthesize comparative insights into their application contexts, including trajectory planning, vision-based navigation, decentralized coordination, and multi-agent conflict resolution, while assessing trade-offs in deployment complexity and operational safety. Beyond method specific analysis, the survey highlights key distinctions, practical challenges, and enabling technologies, concluding with open challenges and future directions for scalable and verifiable UAV swarm intelligence.}, note={Citation Key: KHARGHARIA2026132020}, journal={Neurocomputing}, author={Khargharia, Himadri Sikhar and Ouali, Anis and Shakya, Siddhartha and Ahmad, Sara}, year={2026}, pages={132020} }

% Biography
%\bio{}
% Here goes the biography details.
%\endbio

%\bio{pic1}
% Here goes the biography details.
%\endbio

\end{document}